\documentclass[preprint,3p,times]{elsarticle}

\usepackage[english]{babel}
\usepackage[utf8]{inputenc}
\usepackage{url}
\usepackage{graphicx}
\usepackage{algorithmicx}
\usepackage{algpseudocode}
\algtext*{EndWhile}
\algtext*{EndIf}
\algtext*{EndFor}
\algtext*{EndFunction}
\usepackage{fixltx2e}
\MakeRobust{\Call}
\usepackage{amsmath}
\DeclareMathOperator*{\argmin}{\arg\!\min}
\usepackage{mdframed}
\newmdenv[leftline=false,rightline=false]{topbot}
\usepackage[draft]{todonotes}

\def\operatornamewithlimits{\operatorname*}
\usepackage{tabularx}
\newcolumntype{R}{>{\raggedleft\arraybackslash}X}
\usepackage{siunitx}
\usepackage{booktabs}
\usepackage{rotating}

\journal{Information Sciences}

\begin{document}

\begin{frontmatter}
\title{FRULER: Fuzzy Rule Learning through Evolution for Regression}
\author{I.~Rodr\'iguez-Fdez\corref{irf}}
\ead{ismael.rodriguez@usc.es}
\author{M.~Mucientes\corref{ryc}}
\ead{manuel.mucientes@usc.es}
\author{A.~Bugar\'in\corref{}}
\ead{alberto.bugarin.diz@usc.es}

\cortext[irf]{Corresponding author. Tel.: +34 8818 16392.}

\address{Centro de Investigaci\'on en Tecnolox\'ias da Informaci\'on (CITIUS), Universidade de Santiago de Compostela, SPAIN}

\begin{abstract}
In regression problems, the use of TSK fuzzy systems is widely extended due to the precision of the obtained models.
Moreover, the use of simple linear TSK models is a good choice in many real problems due to the easy understanding of the relationship between the output and input variables. 
In this paper we present FRULER, a new genetic fuzzy system for automatically learning accurate and simple linguistic TSK fuzzy rule bases for regression problems.
In order to reduce the complexity of the learned models while keeping a high accuracy, the algorithm consists of three stages: instance selection, multi-granularity fuzzy discretization of the input variables, and the evolutionary learning of the rule base that uses the Elastic Net regularization to obtain the consequents of the rules.
Each stage was validated using 28 real-world datasets and FRULER was compared with three state of the art genetic fuzzy systems.
Experimental results show that FRULER achieves the most accurate and simple models compared even with approximative approaches.
\end{abstract}

\end{frontmatter}

\section{Introduction}
Predictive modelling aims to build models that use the values of the input variables to predict the expected output.
These models usually have two complementary requirements: accuracy and interpretability \cite{hastie2009elements}.
On one hand, accuracy indicates the ability of the model to predict values close to the real ones. 
On the other hand, interpretability refers to the capability of the model to be understood by a human being \cite{alonso2011special}.
Building models by means of fuzzy rule-based systems combines the interpretability and expressiveness of the rules with the ability of fuzzy logic for representing uncertainty. 
Interpretability for fuzzy systems involves two main issues \cite{alonso2011special}:
\begin{itemize}
\item Readability: it is related with the simplicity of the fuzzy system structure, i.e., the number of variables, linguistic terms per variable, fuzzy rules, premises per rule, etc.
It represents the quantitative or objective part of the interpretability of the model.
\item Comprehensibility: it is determined by the general semantics of the fuzzy system and the fuzzy inference mechanism. It is associated with the fuzzy partitioning of the variables and its meaning for the user, thus representing the qualitative or subjective part of the interpretability.
\end{itemize}

The most important aspect of a fuzzy system in terms of interpretability is the definition of the fuzzy partition for each variable, also called the data base definition.
Two different approaches can be used to define the data base: (i) linguistic, in which all rules share the same fuzzy partition for each variable; (ii) and approximative, which uses a different definition of the fuzzy labels for each rule in the rule base.
The former implies more interpretability through a higher simplicity and comprehensibility, while the latter usually obtains more accurate solutions.
However, approximative approaches can lead to complex partitions of the input space that can make difficult to understand how the input is associated with the response.
Moreover, linguistic partitions, where the summation of the degree of fulfillment for all fuzzy sets is equal to 1 for each value inside a domain, are recognized as the most interpretable fuzzy partitions because they satisfy all semantic constraints (distinguishability, coverage, normality, convexity, etc.) \cite{alonso2011special}.

In the case of regression problems, two different alternatives for fuzzy modelling were used in the literature depending on what the main objective pursued was.
Mamdani fuzzy systems, where both antecedent and consequent are represented by fuzzy sets, were primarily used to obtain interpretable models.
Furthermore, precise fuzzy modelling was mainly developed using Takagi-Sugeno-Kang (TSK) fuzzy knowledge systems \cite{takagi1985fuzzy,sugeno1988structure}, where the antecedents are still represented by fuzzy sets, while the consequent is a weighted combination of the input variables.
Although Mamdani systems are well-known for its semantic interpretability, the linear model in TSK rules is also a good choice since it is straightforward to understand the relationship between the output and input variables.
This is of particular interest in many fields, such as robotics \cite{rodriguez2015learning,Rodriguez-Fdez11_ssci,Mucientes09_ifsa}, medical imaging \cite{Rodriguez-Fdez12_maeb}, industrial estimation \cite{mucientes2009processing} and optimization of processes \cite{vidal2011machine}.

One of the most widely used learning algorithms for automatic building of fuzzy rule bases are Genetic Fuzzy Systems (GFSs) \cite{cordon2001genetic}, i.e., the combination of evolutionary algorithms and fuzzy logic.
Evolutionary algorithms are appropriate for learning fuzzy rules due to their flexibility ---that allows them to codify any part of the fuzzy rule base system---,  and due to their capability to manage the balance between accuracy and simplicity of the model in an effective way.
In particular, recent developments using multi-objective evolutionary fuzzy systems can be found in \cite{alcala2011fast,gacto2014metsk,sanchez2009obtaining,antonelli2013efficient}, where both Mamdani and TSK systems were proposed to solve large-scale regression problems.
Moreover, in \cite{marquez2013efficient} an adaptive fuzzy inference system was proposed to cope with high-dimensional problems.

The simplicity of the models obtained by GFSs for regression has been mostly achieved in the literature through the control of the number of rules and/or the number of labels used in the rule base through a multi-objective approach \cite{fazzolari2013review,ishibuchi2007analysis}.
More recently, the use of instance selection techniques has received more attention in both classification \cite{garcia2012prototype,fazzolari2013study} and regression \cite{rodriguez-fdez2013selection} problems.
This approach faces two problems at once: decreases the complexity for large-scale problems and reduces the overfitting, as the rules can be generated with a part of the training data and the error of the rule base can be estimated with another part (or the whole) training set.
Moreover, when no expert knowledge is available to determine the fuzzy labels, two different approaches can be applied: uniform discretization combined with lateral displacements \cite{alcala2007genetic}, or non-uniform discretization \cite{ishibuchi2002performance}.
Recently, \cite{fazzolari2014discretization,garcia2013survey} have shown the application of non-uniform discretization techniques to classification problems.

The use of TSK fuzzy rule bases implies another complexity dimension: the polynomial in the consequent ---usually with degree 1 (TSK-1) or 0 (TSK-0).
The most widely used approach for learning the coefficients of the polynomial is the least squares method. 
However, that choice often leads to models that overfit the training data and misbehave in test.
This problem can be solved by shrinking (Ridge regularization) or setting some coefficients to zero (Lasso regularization), obtaining simpler models.
Moreover, a combination of both regularizations, called Elastic Net \cite{zou2005regularization} can be used.

In this paper we present FRULER  (Fuzzy RUle Learning through Evolution for Regression), a new GFS algorithm for obtaining accurate and simple linguistic TSK-1 fuzzy rule base models to solve regression problems.
The simplicity of the fuzzy system aims to improve the readability of the model ---and, therefore, the interpretability--- by obtaining linguistic fuzzy partitions with few labels, a low number of rules, and the regularization of the consequents ---which reduces the number of input variables that contribute to the output.
The main contributions of this work are: 
i) a new instance selection method for regression,
ii) a novel multi-granularity fuzzy discretization of the input variables, in order to obtain non-uniform fuzzy partitions with different degrees of granularity, 
iii) an evolutionary algorithm that uses a fast and scalable method with Elastic Net regularization to generate accurate and simple TSK-1 fuzzy rules.

This work is structured as follows. Section \ref{sec:tsk} defines the TSK model used in this work. Section \ref{sec:method} describes the different stages of the GFS: the instance selection method, the discretization approach, and the evolutionary algorithm. Sec. \ref{sec:results} shows the results of the approach in 28 regression problems, and the comparison with other proposals through statistical tests. Finally, the conclusions are presented in Sec. \ref{sec:conclusion}.

\section{Takagi-Sugeno-Kang fuzzy rule systems\label{sec:tsk}}
Takagi, Sugeno, and Kang proposed in \cite{takagi1985fuzzy,sugeno1988structure} a fuzzy rule model in which the antecedents are comprised of linguistic variables, as in the case of Mamdani \cite{mamdani1974application,mamdani1975experiment}, but the consequent is represented as a polynomial function of the input variables.
These type of rules are called TSK fuzzy rules.
The most common function for the consequent of a TSK rule is a linear combination of the input variables (TSK-1), and its structure is as follows:
\begin{align}\label{eq:tskrule}
    & \text{If $X_1$ is $A_1$ and $X_2$ is $A_2$ and $\dots$ and $X_p$ is $A_p$ then} \nonumber\\
    & \text{$Y = \beta_0 + X_1 \cdot \beta_1 + X_2 \cdot \beta_2 + \dots + X_p \cdot \beta_p$}
\end{align}
where $X_j$ represents the $j$-th input variable, $p$ the number of input variables, $A_j$ is the linguistic fuzzy term for $X_j$, $Y$ is the output variable, and $\beta_j$ is the coefficient associated with $X_j$ in the consequent part of the rule.

The matching degree $h$ between the antecedent of the rule $r_k$ and the current inputs to the system $(x_1, x_2, \dots, x_p)$ is calculated as:
\begin{equation}\label{eq:tskdegree}
h_k = T(A_1^k(x_1), A_2^k(x_2), \dots, A_p^k(x_p))
\end{equation}
where $A_j^k$ is the linguistic fuzzy term for the $j$-th input variable in the $k$-th rule and $T$ is the t-norm conjunctive operator, usually the minimum function. The final output of a TSK fuzzy rule base system composed by $m$ TSK fuzzy rules is computed as the average of the individual rule outputs $Y_k$ weighted by the matching degree:
\begin{equation}\label{eq:tskoutput}
\hat{y} = \frac{\sum_{k=1}^m h_k \cdot Y_k}{\sum_{k=1}^m h_k}
\end{equation}

Linguistic TSK fuzzy rule systems represent a good trade-off between accuracy and interpretability:
\begin{itemize}
\item The use of linguistic terms in the antecedent of the rules provides a full description of the input space due to the shared definition of the fuzzy partitions in the data base of the system.
\item The linear representation of the output allows to obtain accurate solutions using different well-studied statistical methods.
\item The consequent of the rules represented by a linear combination of the input variables allows an easy understanding of the relationship between the inputs and the output.
\end{itemize}
Thus, even if the TSK fuzzy rule systems are less comprehensible in natural language terms than a Mamdani approach, the system can provide useful and understandable information, and is the preferable choice in some domains.
In this article we focus on developing simple and accurate TSK fuzzy rule models based on a linguistic representation of the antecedents.

\section{FRULER description\label{sec:method}}
\begin{figure}[htb!]
\centering
\includegraphics[width=1\columnwidth]{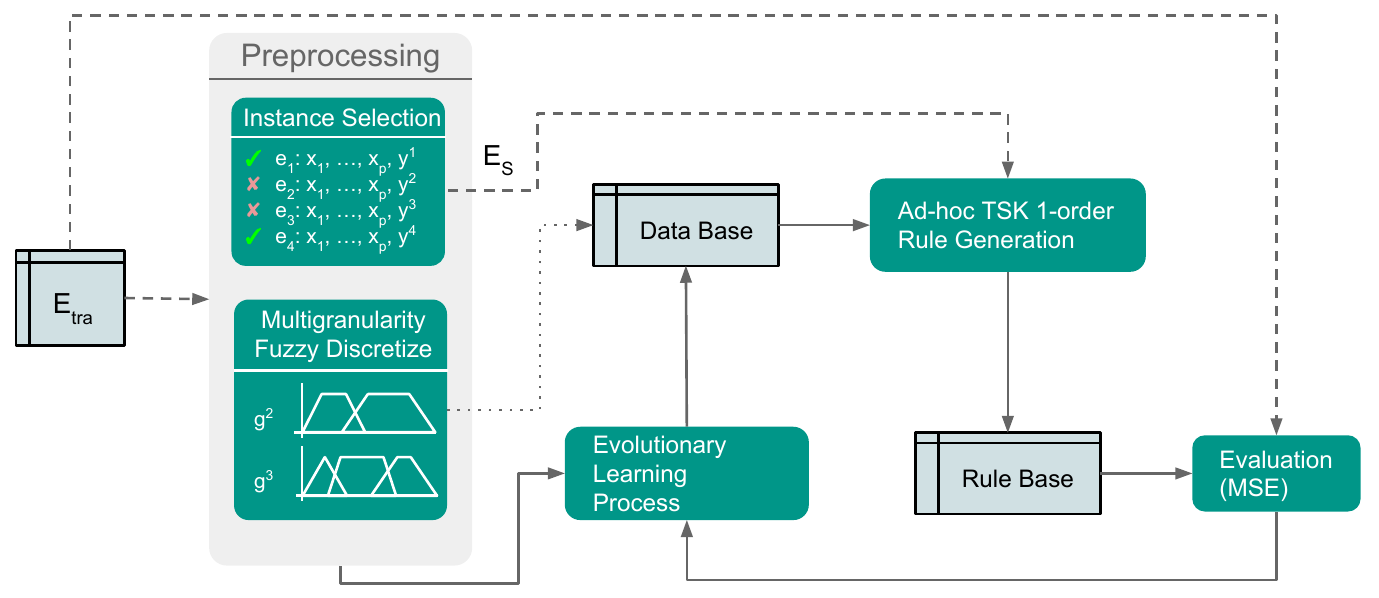}
\caption{FRULER architecture showing each of the three separated stages. Dashed lines indicate flow of data sets, dotted lines multigranularity information and solid lines represent process flow.\label{fig:fruler}}
\end{figure}

This section presents the three main components of FRULER: a two-stage preprocessing ---formed by the instance selection and multi-granularity fuzzy discretization modules---, and a genetic algorithm, which contains an ad-hoc TSK 1-order rule generation module (Fig. \ref{fig:fruler}). 
Both preprocessing techniques are executed to improve the simplicity of the fuzzy rule bases obtained by the evolutionary algorithm. 
On one hand, the instance selection reduces the variance of the models focusing the generated rules on the representative examples.
On the other hand, the multi-granularity fuzzy discretization decreases the complexity of the fuzzy partitions and, therefore, it is not necessary to establish an upper bound in the number of rules in the evolutionary stage.

The evolutionary learning process obtains a definition of the data base of the knowledge system.
Then a novel ad-hoc TSK 1-order rule generation module calculates the antecedents and consequents of each possible rule using only the representative examples.
Finally, each knowledge base generated by the evolutionary algorithm is evaluated using the full training dataset.

\subsection{Instance Selection for Regression\label{sec:is}}
The instance selection method for regression is an improvement of the CCISR (Class Conditional Instance Selection for Regression) algorithm \cite{rodriguez-fdez2013selection}, which is an adaptation for regression of the instance selection method for classification CCIS (Class Conditional Instance Selection) \cite{marchiori2010class}.
The main differences between FRULER instance selection process and CCISR are: 
\begin{itemize}
\item The output variable is discretized in order to simplify the generation of the different graphs needed in the process. However, the discretization does not imply that the CCIS process can be used without modification, as the selected instances must assure a good behavior in regression problems.
\item The error measure is based on the $1-nearest\ neighbor$ ($1NN$) approach for regression, thus reducing the complexity of the calculations compared with CCISR, which uses an ad-hoc fuzzy system to evaluate the instances.
\item The stopping criteria is more flexible, allowing more iterations without improvement until the termination of the process.
\item The size of the initial set of selected examples was also modified, taking the previous improvements into account.
\end{itemize}

The instance selection process is based on a relation called class conditional nearest neighbor (\textit{ccnn}) \cite{marchiori2010class}, defined on pairs of points from a labeled training set as follows: for a given class $c$, $ccnn$ associates to instance $a$ its $nearest\ neighbor$ computed among only those instances (excluded $a$) in class $c$. 
Thus, this relation describes proximity information conditioned to a class label. 

In regression problems, the outputs are real values instead of labels and, therefore, they must be discretized in order to use the \textit{ccnn} relation. 
Traditionally, an unsupervised discretization process needs the definition of either the number of intervals or their shape \cite{dougherty1995supervised}. 
In FRULER, the shape of the intervals is guided by the output density, i.e., the intervals are selected such that they represent dense clusters.
In other words, the split points between intervals are selected in the zones where the density of the output is locally minimum.

We use Kernel Density Estimation (KDE) with a gaussian kernel in order to estimate the probability density function of the output variable ($y$) in a non-parametric way. 
In order to select the appropriate kernel bandwidth, Scott's rule is applied. \cite{scott2009multivariate}.
Once the probability density function is obtained, the local minimum determines the split points, and, therefore, which labels/classes are used for the \textit{ccnn} relation.
Thus, each instance is associated with one of the labels obtained by this process and the instance selection method can follow the CCIS procedure.

Two different graphs can be constructed using this relation, as proposed in CCIS:
\begin{itemize}
 \item Within-class directed graph ($G_{wc}$): consists in a graph where each instance has an edge pointing to the nearest instance of the same class.
 \item Between-class directed graph ($G_{bc}$): is a graph where each instance has an edge pointing to the nearest instance of any different class.
\end{itemize}

These graphs are used to define an instance scoring function by means of a directed information-theoretic measure (the K-divergence) applied to the in-degree distributions of these graphs. 
The scoring function, named \texttt{Score}, is defined as: 
\begin{equation}
    \textit{Score}(e^i) = p^i_w \cdot log \left(\frac{p^i_w}{(p^i_w + p^i_b)/2}\right) - p^i_b \cdot log \left(\frac{p^i_b}{(p^i_w + p^i_b)/2}\right)
\end{equation}
where $e^i$ is the example to which the score is calculated, $p^i_w$ is the in-degree of $e^i$ in $G_{wc}$ divided by the total in-degree of $G_{wc}$, and $p^i_b$ is the inner degree of $e^i$ in $G_{bc}$ divided by the total in-degree of $G_{bc}$.
This scoring function is used to develop an effective large margin instance selection method, called Class Conditional selection (Fig. \ref{CC}). 

The instance selection algorithm starts from a set of training examples:
\begin{equation}
E = \{e^1,e^2,\dots,e^n\}
\end{equation}
where $n$ is the number of examples.
The method proposed here uses the \textit{leave-one-out} mean squared error (MSE) with $\it{1NN}$ (this error is called $\epsilon$) in order to estimate the information loss.
Thus, although the scoring function and the graphs are based on the labels obtained by KDE, the error measure is based on the original regression problem.

\begin{figure}[htb!]
\begin{topbot}
\begin{minipage}[c]{\columnwidth}
\begin{algorithmic}[1]
\State $\{e^1,\dots,e^n\} = E$ sorted in decreasing order of \texttt{Score}
\State $S=\{e^1,\dots,e^{k_0}\}$
\State $it_{wi} = 0$
\Repeat
  \State $\it{Temp} = S \cup \{e^l\}$
  \If{$\epsilon^{\it{Temp}} < \epsilon^S$}
    \State $it_{wi} = 0$
    \State $S_{best} = \it{Temp}$
  \Else
    \State $it_{wi} = it_{wi} + 1$
  \EndIf
  \State $S = \it{Temp}$
\Until{$E = S \vee it_{wi} > \sqrt{|E|/|S|} $}

\State return $S_{best}$
\end{algorithmic}
\end{minipage}
\end{topbot}
\caption{Pseudocode of Class Conditional selection \cite{marchiori2010class}.}
\label{CC}
\end{figure}

First, an initial core of instances from $E$ is selected, sorted by \texttt{Score} (Fig. \ref{CC}, lines 1-2). The size of this initial set is:
\begin{equation}
    k_0 = max\left(c,\left\lceil \frac{\epsilon^{E} \cdot |E|}{max(y) - min(y)} \right\rceil\right)
\end{equation}
where $c$ is the number of classes obtained from KDE and $\epsilon^{E}$ is the error using the set of examples in $E$. 
This choice is motivated  because (i) there has to be at least one example for each class, and (ii) the error in the second part can be interpreted as the miss-classification probability divided by the range of the output $max(y) - min(y)$.
Thus, the second part indicates that at least the miss-classified examples must be selected in order to be correctly classified.
After this, the instance selection method iteratively selects instances and adds them to the set $S$ (lines 4-12), choosing in first place those with the highest score. 
The process terminates when all the examples of $E$ are in $S$ or when $it_{wi}$ ---the number of consecutive iterations for which the empirical error ($\epsilon^S$) increases--- is greater than $\sqrt{|E|/|S|}$ (line 12).
This threshold allows more iterations without improvement at the beginning of the selection process, when the error is more sensitive, and stops earlier when the number of selected instances is high.

In order to further improve the number of selected instances, CCIS uses the Thin-out post-processing (Fig. \ref{THIN}). This algorithm selects points close to the decision boundary of the $\it{1NN}$ rule. This is achieved by selecting instances having positive in-degree in the between-class graph set $S$ ($G_{bc}^S$) and storing them in $S_f$. Then, an iterative process is done as follows: points having positive in-degree in the $G_{bc}^{S_1}$ are added to $S_f$ if they were not “isolated” in the previous iteration, that is, if their in-degree was not zero (line 6). This iterative process terminates when the empirical error increases (line 7).

\begin{figure}[htb!]
\begin{topbot} \begin{minipage}[c]{\columnwidth}
\begin{algorithmic}[1]
\State $S_f = \{e^l \in S$ with in-degree in $G_{bc}^S > 0\}$
\State $S_{prev} = S$
\State $S_1 = S \setminus S_f$
\State $go\_on = true$
\While{$go\_on$}
  \State $S_t = \{e \in S_1$ with in-degree in $G_{bc}^{S_1} > 0$ and with in-degree in $G_{bc}^{S_{prev}} > 0\}$
  \State $go\_on = \epsilon^{S_f \cup S_t} < \epsilon^{S_f}$
  \If{$go\_on$}
    \State $S_f = S_f \cup S_t$
    \State $S_{prev} = S_1$
    \State $S_1 = S \setminus S_f$
  \EndIf
\EndWhile

\State return $S_f$
\end{algorithmic}
\end{minipage} \end{topbot}
\caption{Pseudocode of Thin-out selection \cite{marchiori2010class}.}
\label{THIN}
\end{figure}

\subsection{Multi-granularity Fuzzy Discretization for Regression\label{sec:fd}}
The definition of the fuzzy partition of each input variable is a critical step in the design of TSK fuzzy rule bases.
When no knowledge is available, the set of fuzzy labels for a variable is automatically obtained through fuzzy discretization.
Moreover, if the number of labels is unknown, then a multi-granularity approach is used, i.e., the definition of a different fuzzy partition for each regarded granularity. 
Specifically, a granularity $g^i_{var}$ divides the variable $var$ in $i$ fuzzy labels, i.e., $g^i_{var} = \{A^{i,1}_{var}, \dots, A^{i,i}_{var}\}$.

\begin{figure}[htb!]
\centering
\includegraphics[width=0.6\columnwidth]{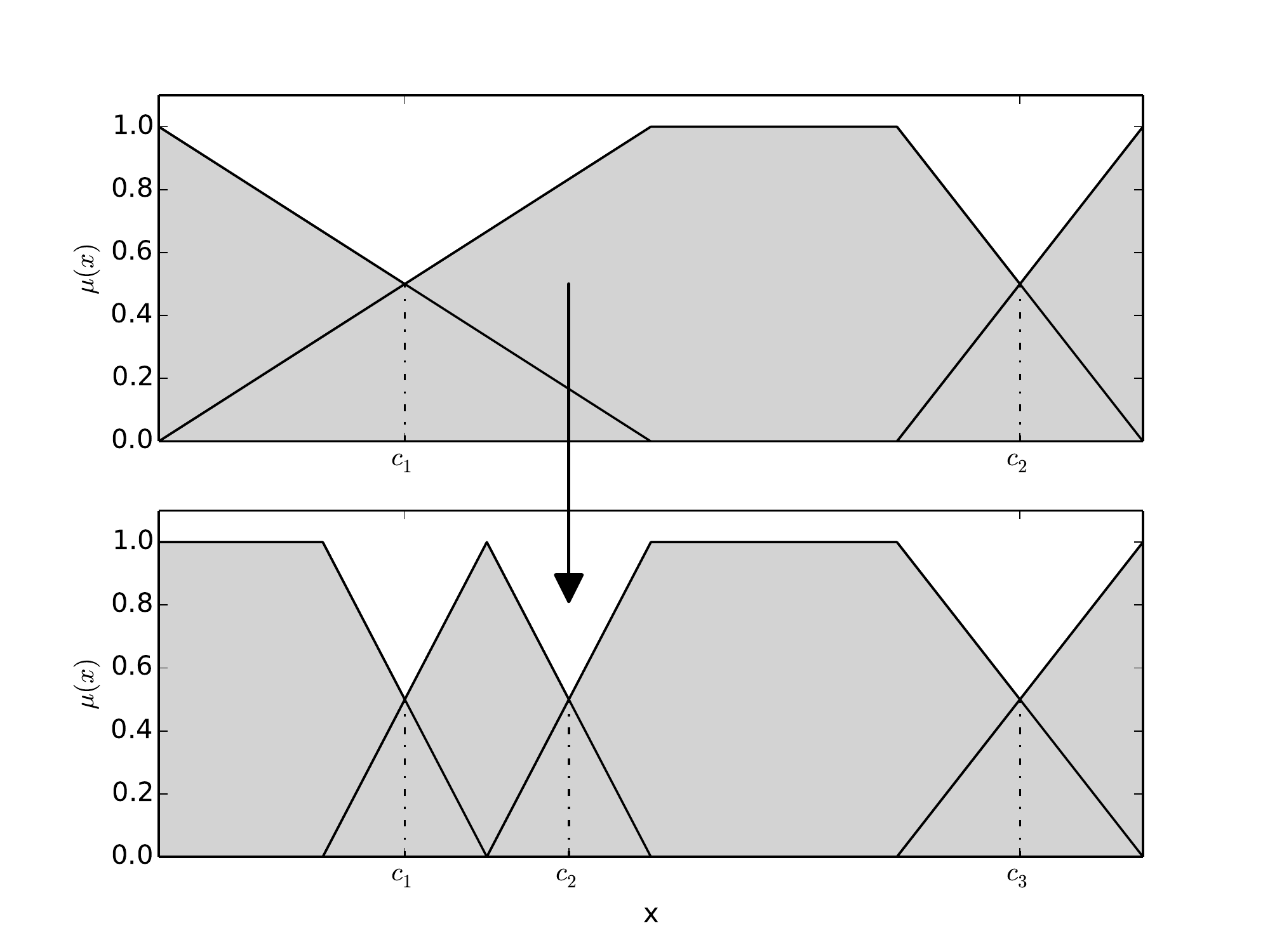}
\caption{Top-down approach for the multi-granularity discretization. Only one label is divided into two new labels in order to obtain the next granularity.\label{fig:granularities}}
\end{figure}

The generation of the fuzzy linguistic labels can be divided into two stages. First, the variable must be discretized to obtain a set of split points $C^g$ for each granularity $g$. Then, given the split points, the fuzzy labels can be defined for each granularity.
In a top-down approach, the split points are searched iteratively, i.e., only a new split point is added at each step, obtaining two new intervals.
Therefore, the approach proposed in this work aims to preserve interpretability between contiguous granularities: adding a new label to the previous granularity and modifying the flanks of the adjacent labels (Fig. \ref{fig:granularities}).
In regression problems (TSK-1 in our case), the discretization process must search for the split point that minimizes the error when a linear model is applied to each of the resulting intervals.

In order to select the maximum number of split points for a variable, we have used the well-known Bayesian Information Criterion (BIC).
This measure can be separated into two parts: the error of applying the model to the data and its complexity.
In this case, the error is obtained from the summation of the mean squared error of a least squares fitted model for each interval of the discretization.
On the other hand, the complexity of the model is determined by the number of parameters, in this case the number of inner splits and the parameters fitted by each regression applied in each interval.

The pseudocode of the discretization method for a variable is shown in Fig. \ref{fig:disc}.
First, the split points for granularity $1$ are initialized using the domain limits (line 2).
The BIC measure for this first granularity is calculated (line 3) using $\it{MSE}$, a function that gets a set of examples $X$, learns a linear regression model using least squares and, finally, calculates the mean squared error of the model.
In this case, the number of parameters is two, corresponding to the coefficients of the linear model.
After that, an iterative process is executed: at each step, the split points of a new granularity are defined adding a new split point to the previous granularity (lines 5-16).

\begin{figure}[htb!]
\begin{topbot} \begin{minipage}[c]{\columnwidth}
\begin{algorithmic}[1]
\State $g = 1$
\State $C^g = \{min(X), max(X)\}$
\State $BIC^g = |X| \cdot log(\it{MSE}(X)) + 2\cdot log(|X|)$
\State $it_{wi} = 0$
\Repeat
    \State $C = \big\{c_i\ |\ c_i = \argmin_{c} $ \Call{LinearError}{$\{x \in X : C^g_i < x < C^g_{i+1}\}, c$}, $\forall i = 0,...,g$, $\forall c \in [C^g_i, C^g_{i+1}]\big\}$
    \State $i_{min} = \argmin_i$ \Call{LinearError}{$\{x \in X : C^g_i < x < C^g_{i+1}\}$, $c_i$}, $\forall c_i \in C$
    \State $C^{g+1} = C^{g} \cup \{c_{i_{min}}\}$
    \State $g = g + 1$
    \State $BIC^g = |X| \cdot log(\sum_{i=0}^{g}\it{MSE}(\{x \in X : C^g_i < x < C^g_{i+1}\}) + (|C^g|-2) \cdot 2\cdot log(|X|)$
    \If{$BIC^g < BIC^{min}$}
        \State $it_{wi} = 0$
        \State $min = g$
    \Else
        \State $it_{wi} = it_{wi} +1$
    \EndIf
\Until{$it_{wi} > \sqrt{\frac{|X|}{30}/min}$}
\State \Return $C^1,\dots,C^{min}$
\end{algorithmic}
\end{minipage} \end{topbot}
\caption{Pseudocode of the discretization method.}
\label{fig:disc}
\end{figure}

In order to obtain the split point for the new granularity, first, the best split point ($c_i$) for each interval between the split points of the previous granularity ($[C^g_i, G^g_{i+1}]$) is obtained (line 6).
The best split point is defined as the point that obtains the global minimum of the function \texttt{LinearError} (Fig. \ref{fig:linearerror}) in an interval (Fig. \ref{fig:disc}, line 7).
\texttt{LinearError} gets a set of examples $X$ and a split point $c$ and calculates the total squared error ($\it{SE}$) of $X$, which is calculated with the corresponding linear regression models at each side of the split point.
Only split points that obtain intervals with size of at least $30$ are taken into account to assure that the obtained linear regressions are statistically valid.

The selected split point is added to the new granularity split points (lines 8-9), and the BIC measure is calculated (line 10).
The number of parameters used for the BIC measure is $2$ (coefficients of the linear regression for a single variable) for each interval.
The number of intervals is calculated as $|C^g|-2$, where $2$ is subtracted to disregard the split points at the end of the domain of variable $X$.

\begin{figure}[htb!]
\begin{topbot} \begin{minipage}[c]{\columnwidth}
\begin{algorithmic}[1]
\Function{LinearError}{$X$, $c$}
    \State $X_l = \{x \in X: x < c\}$
    \State $X_r = \{x \in X: x > c\}$
    \State \Return $\it{SE}(X_l)\cdot\frac{|X_l|}{|X|} + SE(X_r)\cdot\frac{|X_r|}{|X|}$
\EndFunction
\end{algorithmic}
\end{minipage} \end{topbot}
\caption{Pseudocode of the function to be minimized by the discretization method.}
\label{fig:linearerror}
\end{figure}

Finally, when the number of consecutive iterations without improvement in the BIC value ($it_{wi}$) is greater than $\sqrt{\frac{|X|}{30}/min}$, the algorithm stops (lines 11-16).
This criterion ensures that at the beginning of the discretization process ---the granularity is low---, the BIC may worsen for more iterations, while with larger granularities, the algorithm becomes stricter in the stopping criterion.
The number of data points is divided by $30$ in order to obtain the maximum number of intervals.

After obtaining the discretization of the variable for each granularity, the method proposed in \cite{ishibuchi2002performance} is applied for each $C^g$ ---set of split points for the granularity $g$--- in order to get the multi-granularity fuzzy partitions.
This method uses a fuzziness parameter that indicates how fuzzy are the linguistic labels. 
A fuzziness 0 indicates crisp intervals, while a fuzziness 1 indicates the selection of a fuzzy set with the smallest kernel ---set of points with membership equal to 1.

\subsection{Evolutionary Algorithm\label{sec:ea}}
The evolutionary algorithm learns a linguistic TSK model.
The integration of the evolutionary algorithm with the preprocessing stage is as follows (Fig. \ref{fig:fruler}):
\begin{itemize}
\item First, the instance selection process is executed over the training examples $E_{tra}$ in order to obtain a subset of representative examples $E_{S}$.
\item Then, the multi-granularity fuzzy discretization process obtains the fuzzy partitions for each input variable.
\item Finally, the evolutionary algorithm searches for the best data base configuration using the obtained fuzzy partitions, generates the entire linguistic TSK rule base using $E_{S}$ and evaluates the different rule bases using $E_{tra}$.
\end{itemize}

\begin{figure}[htb!]
\centering
\includegraphics[width=1\columnwidth]{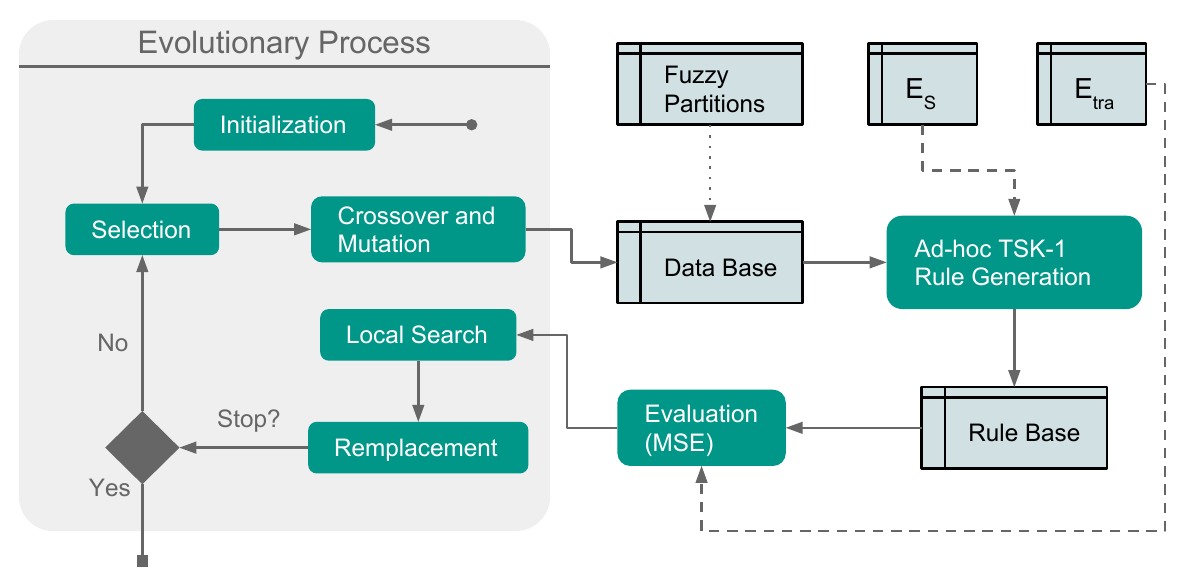}
\caption{The Evolutionary learning process used in FRULER. Dashed lines indicate flow of data sets, dotted lines are for multigranularity information and solid lines represent process flow.\label{fig:evol}}
\end{figure}

Figure \ref{fig:evol} shows the evolutionary learning process and how it uses the fuzzy partitions and the training examples. In what follows, we describe in detail the different components of the evolutionary algorithm.

\subsubsection{Chromosome Codification}
The chromosome codification represents the parameters needed to create the data base and the rule base.
Each individual has to codify a single fuzzy partition for each input variable from the fuzzy partitions obtained in the multi-granularity fuzzy discretization (Sec. \ref{sec:fd}).
Moreover, the individuals also use the 2-tuple representation of the labels \cite{herrera20002}.
This approach applies a displacement of a linguistic term within an interval that expresses the movement of a label between its two adjacent labels.
In our case, a different displacement is going to be applied to each of the split points.

\begin{figure}[htb!]
\centering
\includegraphics[width=\columnwidth]{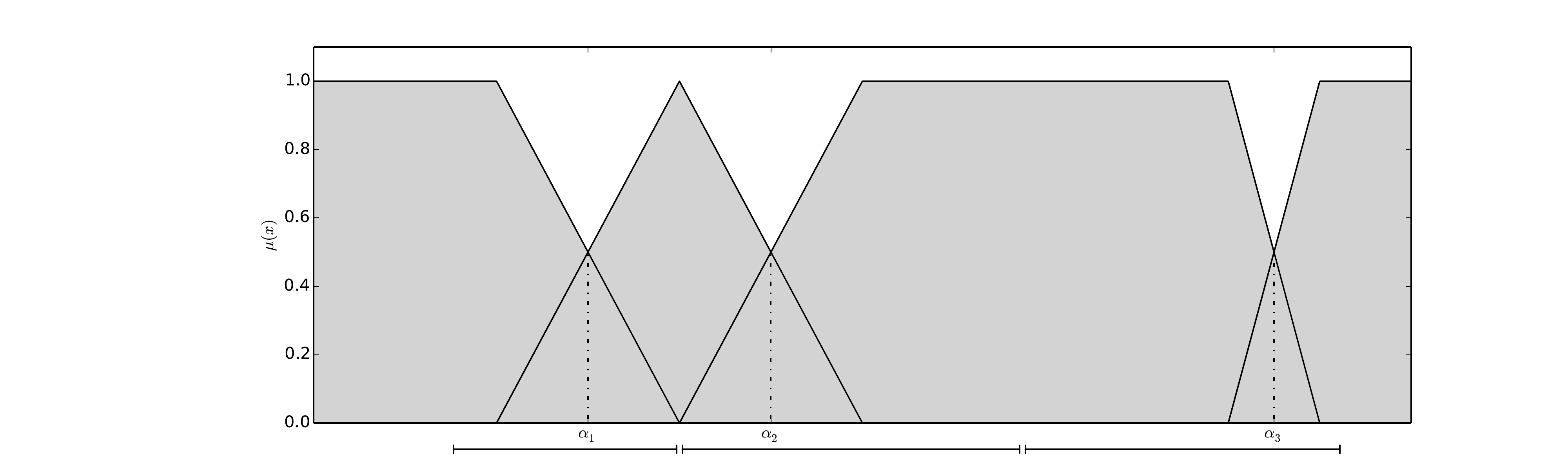}
\caption{An example of lateral displacement intervals for limits equal to $(0.5, 0.5)$. The split points can move a maximum of half of the distance to the next split point.\label{fig:ld_interval}}
\end{figure}

\begin{figure}[htb!]
\centering
\includegraphics[width=\columnwidth]{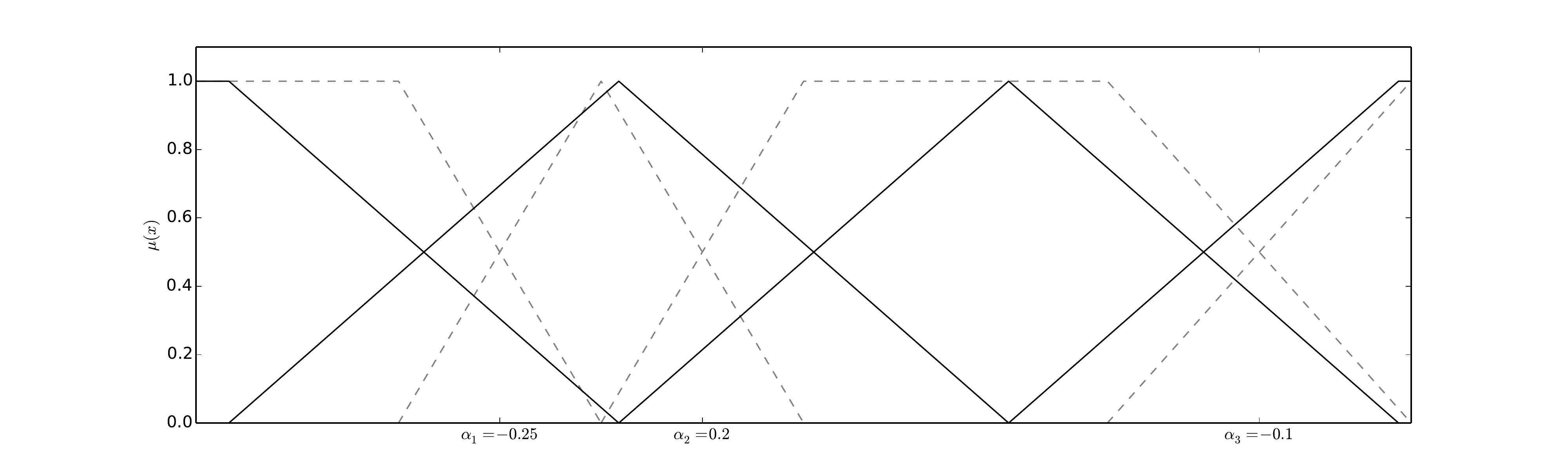}
\caption{A lateral displacement example. The dashed lines indicate the original fuzzy partition, while the solid lines indicate the obtained partition after the displacement has been applied.\label{fig:ld_example}}
\end{figure}

Thus, the chromosome is codified with a double coding scheme ($C = C_1 + C_2$):
\begin{itemize}
\item $C_1$ represents the granularity used in each input variable. It is codified with a vector of $p$ integers:
\begin{equation}
C_1 = (g_1, g_2, \dots, g_p)
\end{equation}
where $g_{i}$ represents the granularity selected for input variable $i$. 
When the granularity of a variable is equal to $1$, then it is not used in the antecedent part.
However, this variable can still be used in the consequent, since it could be relevant for calculating the output.

\item $C_2$ represents the lateral displacements of the split points of the input variables fuzzy partitions. Thus, the length of $C_2$ depends on the granularity for each input variable: $|C_2| = \sum_{j=1}^p{(|g_j|-1)}, \forall g_j \in C_1$:
\begin{equation}
C_2 = (\alpha^1_1, \dots, \alpha^{g_1-1}_1, \dots, \alpha^{1}_p, \dots, \alpha^{g_p-1}_p)
\end{equation}
where $\alpha^j_i$ represents the lateral displacement of the $j$ split point of variable $i$.
Each lateral displacement can vary in the $(0.5, 0.5$) interval which represents half of the distance between each split point (Fig. \ref{fig:ld_interval}).
An example of a lateral displacement can be seen in Figure \ref{fig:ld_example}.
The fuzzy partitions are always strong ---the sum of the degree of fulfillment for each point of the domain is always equal to $1$--- and, therefore, interpretability is maintained.
\end{itemize}

\subsubsection{Initialization}
The initial pool of individuals is generated by a combination of two initialization procedures. 
A half of the individuals are generated with the same random granularity for each variable, while the other half is created with a different random granularity for each variable. The lateral displacements are initialized to 0 in all cases.

After that, when the product of the granularities indicated in $C_1$ (i.e., the maximum number of rules that can be obtained) is greater than the number of input variables times the highest maximum granularity of the variables, then a variable is randomly selected to be removed from the antecedent part ---its granularity is set to 1--- until the previous condition is satisfied.
This is done in order to avoid too complex solutions in the initialization stage ---during the evolutionary learning this upper bound to the number of rules does not apply.

\subsubsection{TSK Rule Base Generation\label{sec:sgd}}
An ad-hoc method is used to construct the rule base from the data base codified in the chromosome, i.e. the fuzzy partitions indicated in $C_1$ after applying the displacement in $C_2$.
The Wang \& Mendel algorithm \cite{wang1992generating} is used to create the antecedent part of the rule base for each individual. 
The method is quick and simple, and obtains a representative rule base given the definition of the data base and a set of examples.

The consequent part of the rules is learned using the Elastic Net method \cite{zou2005regularization} in order to obtain the coefficients of the degree 1 polynomial for each rule.
Elastic Net linearly combines the $\ell_1$ (Lasso regularization) and $\ell_2$ (Ridge regularization penalties of the Lasso and Ridge methods to overcome some of their limitations.
This combination allows obtaining sparse models ---forces variables with little or none correlation with the output to have coefficients equal to 0--- while learning a smooth linear regression ---the coefficients are shrunk towards 0.

Elastic Net obtains the coefficients of a linear regression minimizing the following equation:
\begin{equation}\label{eq:min}
\hat{\beta} = \operatornamewithlimits{arg\,min}_{\beta} ||Y - X \cdot \beta||_2^2 + \lambda \cdot \alpha \cdot \|\beta\|^2_2 + \lambda \cdot (1 - \alpha) \cdot \|\beta\|_1
\end{equation}
where $\beta$ is the coefficients vector $(\beta_0, \beta_1, \dots, \beta_p)$, $Y$ is the outputs vector $(y^1, \dots, y^n)$, $X$ is the inputs matrix with size $n \times p$ ---rows represent examples while columns are the input variables---, $\lambda$ is the regularization parameter and $\alpha$ represents the trade-off between $\ell_1$ and $\ell_2$ penalization. 

In order to use Elastic Net for learning the consequents, the coefficients for each rule cannot be calculated separately due to the aggregation function used to obtain the output of the system (Eq. \ref{eq:tskoutput}). 
Therefore, all the coefficients must be optimized at the same time, taking into account the degree of fulfillment of each rule (Eq. \ref{eq:tskdegree}) for each input vector $x^i$.
Thus, the matrix $X$ is modified as follows:
\begin{itemize}
\item The normalized degree of fulfillment for each rule $r_k$ for each example $e^i$ is calculated as:
\begin{equation}
z_k^i = \frac{h_k(x^i)}{\sum_{u=1}^m h_u(x^i)}
\end{equation}
where the denominator is the normalization term for each input vector $x^i$, i.e., the summation of the degree of fulfillment of all rules.
\item Then, the matrix $X$ is defined as: 
\begin{equation}\label{eq:x}
X = 
\begin{pmatrix}
z^1_1, & x^1_1 \cdot z^1_1, & \dots, & x^1_p \cdot z^1_1, & \dots, & z^1_m, & x^1_1 \cdot z^1_m, & \dots, & x^1_p \cdot z^1_m \\
\vdots & & & & & & & & \vdots \\
z^n_1, & x^n_1 \cdot z^n_1, & \dots, & x^n_p \cdot z^n_1, & \dots, & z^n_m, & x^n_1 \cdot z^n_m, & \dots, & x^n_p \cdot z^n_m \\
\end{pmatrix}
\end{equation}
where each row replicates the input vector $x^i = (1, x^i_1, x^i_2, \dots, x^i_p)$ ---where a 1 was added to take into account the independent term--- as many times as the number of rules ($m$), weighting each rule $r_k$ by $z^i_k$.
\item Finally, the coefficient vector is the concatenation of the coefficients of all rules:
\begin{equation}
\beta = 
\begin{pmatrix}
\beta^1_0, & \beta^1_1, & \dots, & \beta^1_p, & \dots, & \beta^m_0, & \beta^m_1, & \dots, & \beta^m_p \\
\end{pmatrix}
\end{equation}
\end{itemize}

In order to solve the minimization problem of Elastic Net (Eq. \ref{eq:min}), the Stochastic Gradient Descent (SGD) optimization technique was used \cite{bottou2010large,tsuruoka2009stochastic}.
This gradient descent method is characterized by updating each coefficient separately using only one example at a time.
This is particularly suited for sparse datasets, which is a common case when $X$ is constructed using Eq. \ref{eq:x} ---$z^i_k$ is $0$ when a rule does not cover an example.

The pseudocode of the method is shown in Figure \ref{fig:sgd}. SGD needs three parameters to solve the Elastic Net approach: the regularization ($\lambda$), the trade-off between Lasso and Ridge ($\alpha$) and the initial learning rate ($\eta^0$). 
On one hand, $\alpha$ usually takes a low value in order to behave like $\ell_1$ but with the shrinkage of $\ell_2$ in the features with coefficient not equal to 0.
On the other hand, $\lambda$ and $\eta^0$ can be obtained using a grid search ---testing a set of possible values between a predefined interval--- using only a small subset of examples since the convergence properties are maintained \cite{bottou2010large}.

The algorithm is composed of three different loops: i) lines 3-27 which represent an iteration over the whole dataset, ii) lines 6-17 which iterate over each example and iii) lines 11-17 which iterate over the coefficients.
Note that, in this case, the number of coefficients is the number of columns in $X$ ($p \cdot m$), i.e., the number of input variables of the problem ($p$) times the number of rules ($m$).
First, the examples are shuffled (line 5) each time the whole dataset is used.
Then, for each example $e^i$, $t$ is incremented by 1 and the learning rate ($\eta^t$) and the shrinkage portion for both $\ell_1$ and $\ell_2$ ($s$ and $u$ respectively) are updated (lines 6-10).
After that, for each coefficient $w_j$, line 12 applies Ridge regularization \cite{bottou2010large}, while lines 13-17 apply the Lasso approach \cite{tsuruoka2009stochastic}.
The Lasso approach uses thresholds in order to decide if the variable is going to be selected (weight different from 0) and updates the threshold for each input variable for the next iterations ($q_j$ in line 17).
Finally, the coefficient of determination $R^2$ is calculated (line 18) and compared with the best obtained so far.
If it is better, then the estimated coefficients $\hat{\beta}$ are updated and, if it is not, the number of iterations without improvement ($it_{wi}$) is incremented by 1.
When $it_{wi}$ exceeds the threshold defined in line 27, the algorithm stops.
This threshold is directly proportional to the number of examples, and decreases with the number of iterations.

\begin{figure}[htb!]
\begin{topbot} \begin{minipage}[c]{\columnwidth}
\begin{algorithmic}[1]
\Function{SGD-ElasticNet}{X, Y, $\lambda$, $\alpha$, $\eta^0$}
    \State $it = 0,\ t = 0,\ s = 1,\ u = 0,\ w^0 = 0_{1\times p},\ q = 0_{1\times p} $
    \Repeat
        \State \Call{Shuffle\_Rows}{X} \Comment{SGD needs to reorder the rows of $X$}
        \For{$i = 1,\dots,n$}
            \State $t = t + 1$ \Comment{$t$ counts how many updates of the weights have been applied}
            \State $\eta^t = \eta^0 \cdot (\lambda \cdot t)^{-1}$ \Comment{Updates the learning rate to be more conservative}
            \State $\hat{y}^i = x^i \cdot w^t \cdot s$ \Comment{Obtains the estimated output}
            \State $s = s \cdot (1 - \alpha \cdot \eta^t \cdot \lambda)$ \Comment{Updates how much of $\ell_2$ was applied}
            \State $u = u + (1 - \alpha) \cdot \eta^t \cdot \lambda$ \Comment{Updates how much of $\ell_1$ was applied}
            \For{$j = 1,\dots,p$}
                \State $w_j^{t+\frac{1}{2}} = w_j^{t} - \eta^t \cdot (\hat{y}^i - y^i) \cdot x^i_j / s$ \Comment{Applies the $\ell_2$ regularization}
                \If{$s \cdot w_j^{t+\frac{1}{2}} > 0$} \Comment{Applies $\ell_1$ regularization and thresholds in the following 5 lines}
                    \State $w_j^{t+1} = \textnormal{max}(0, w_j^{t+\frac{1}{2}} - (u + q_j) / s)$ \Comment{Positive threshold} 
                \ElsIf{$s \cdot w_j^{t+\frac{1}{2}} < 0$} 
                    \State $w_j^{t+1} = \textnormal{min}(0, w_j^{t+\frac{1}{2}} + (u - q_j) / s)$ \Comment{Negative threshold}
                \EndIf
                \State $q_j = q_j + s \cdot (w_j^{t+1} - w_j^{t+\frac{1}{2}})$ \Comment{Updates thresholds}
            \EndFor
        \EndFor
        \State $R^2_it = 1 - \frac{1}{n} \sum_{i=0}^{n} (x^i \cdot w^{t+1} \cdot s - y^i)^2$ \Comment{Calculates the coefficient of determination}
        \If{$R^2_it > R^2_{best}$} \Comment{Updates the best values so far and the iterations without improvement}
            \State $\hat{\beta} = w^{t+1} \cdot s$
            \State $R^2_{best} = R^2_t$
            \State $it_{wi} = 0$
        \Else
            \State $it_{wi} = it_{wi} + 1$
        \EndIf
        \State $it = it + 1$ \Comment{$it$ counts how many times the full dataset was used}
    \Until{$it_{wi} >  sqrt(|X|/it)$}
\State \Return $\hat{\beta}$
\EndFunction
\end{algorithmic}
\end{minipage} \end{topbot}
\caption{Pseudocode of SGD for Elastic-Net.}
\label{fig:sgd}
\end{figure}

Only those examples in $E_{s}$ are used to obtain the rule base from the codified chromosome. 
In this manner, those examples that are not representative are not considered for the rule generation. 
Thus, the method avoids the creation of too specific rules, and reduces the time needed to create the rule base.

\subsubsection{Evaluation}
The fitness function is based on the estimation of the error of the generated rule base:
\begin{equation}\label{eq:mse}
\it{fitness} = \it{MSE(E_{tra})} = \frac{1}{2\cdot|E|}\sum_{i=1}^{|E|} (F(x^i) - y^i)^2,
\end{equation}
where $E_{tra}$ is the full training dataset and $F(x^i)$ is the output obtained by the knowledge base for the input $x^i$. 
Using all the examples for evaluation can be seen, in some way, as a validation process, as the rule base was constructed with a subset of them ($E_{S}$).

\subsubsection{Selection and Replacement}
The selection is performed by a binary tournament.
On the other hand, the replacement method joins the previous and current populations, and selects the $N$ best individuals as the new population. 

\subsubsection{Crossover and Mutation}
Two crossover operations are defined: one-point crossover for exchanging the $C_1$ parts (it also exchanges the corresponding $C_2$ genes) and, when the $C_1$ parts are equal, the parent-centric BLX (PCBLX) \cite{herrera2003taxonomy} is used to crossover the $C_2$ part.
In order to prevent the crossover of too similar individuals, an incest prevention was implemented.
When the euclidean distance of the lateral displacements is less than a particular threshold $L$, the individuals are not crossed.

The mutation (with probability $p_{mut}$) applies two possible operations with equal probability to a randomly selected gene of the $C_1$ part: i) decreasing the granularity by $1$ or ii) increasing the granularity to a more specific granularity ---all the granularities have the same chance.
In order to calculate the new lateral displacements in the corresponding $C_2$ part, the displacements of the previous granularity are taken into account.
The displacement associated with a particular split point is calculated adding the displacements of the two nearest split points of the previous granularity (before mutation) weighted by the distance between the split points.

\subsubsection{Local Search}
After the replacement, all the new individuals (their $C_1$ part of the chromosome was not generated before) are used in a local search process.
This stage generates $n_{ls}$ new $C_1$ parts with equal or less granularity ---with equal probability--- for each variable.
Then, the $C_2$ part is generated randomly with a uniform distribution in the $(-0.5, 0.5)$ interval.
The new chromosomes are decoded and evaluated and, if there is a solution that obtains better fitness, then it replaces the original individual.

\subsubsection{Restart and Stopping Criteria}
The restart mechanism uses the incest prevention threshold $L$ as a trigger.
First, $L$ is initialized as the maximum length of the $C_2$ part, i.e. the product of the number of input variables times the largest maximum granularity of the variables, divided by $4$.
This implies that the incest prevention allows crossovers between individuals that have a distance higher than a quarter of the maximum euclidean distance.
Then, for each iteration, $L$ is decreased in different ways in order to accelerate convergence:
\begin{itemize}
\item $L$ is decreased by $0.4$ in all the iterations, in order to increase convergence.
\item If there are no new individuals in the population, then $L$ is decreased by $0.2$.
\item If the best individual does not change, $L$ is also decreased by $0.2$.
\end{itemize}
Finally, when $L$ reaches $0$, the population is restarted, and $L$ is reinitialized.
Only the best individual so far is kept, and the local search process is executed to the best individual in order to generate new individuals until the population is complete.
When the restart criterion is fulfilled twice, the algorithm stops, i.e., one single restart is executed.
Moreover, if the number of evaluations reaches a threshold, then the algorithm is also stopped.
When the evolutionary algorithm stops, the best rule base consequents are optimized applying the SGD algorithm (Sec. \ref{sec:sgd}) using all the training examples.

\section{Results\label{sec:results}}
In order to analyze the performance of FRULER, we have used 28 real-world regression problems from the KEEL project repository \cite{alcala2009keel}. 
Table \ref{tab:datasets} shows the characteristics of the datasets, with the number of instances ranging from 337 to 40,768 examples, and the number of input variables from 2 to 40.
The most complex problems ---large scale--- due to both the number of examples and variables are the ones in the last 8 rows (Table \ref{tab:datasets}).

\begin{table}[htb!]
\centering
    \begin{tabular}{llrS[table-format = 5.0, input-decimal-markers = .,input-ignore = {,},table-number-alignment = right,
  group-separator={,}, group-four-digits = true]}
    \toprule
    Problem                      & Abbr.  & {\# Variables} & {\# Cases} \\
    \midrule
    Electrical Length             & ELE1   & 2         & 495   \\
    Plastic Strength              & PLA    & 2         & 1650  \\
    Quake                         & QUA    & 3         & 2178  \\
    Electrical Maintenance        & ELE2   & 4         & 1056  \\
    Friedman                      & FRIE   & 5         & 1200  \\
    Auto MPG6                     & MPG6   & 5         & 398   \\
    Delta Ailerons                & DELAIL & 5         & 7129  \\
    Daily Electricity Energy      & DEE    & 6         & 365   \\
    Delta Elevators               & DELELV & 6         & 9517  \\
    Analcat                       & ANA    & 7         & 4052  \\
    Auto MPG8                     & MPG8   & 7         & 398   \\
    Abalone                       & ABA    & 8         & 4177  \\
    Concrete Compressive Strength & CON    & 8         & 1030  \\
    Stock prices                  & STP    & 9         & 950   \\
    Weather Ankara                & WAN    & 9         & 1609  \\
    Weather Izmir                 & WIZ    & 9         & 1461  \\
    Forest Fires                  & FOR    & 12        & 517   \\
    Mortgage                      & MOR    & 15        & 1049  \\
    Treasury                      & TRE    & 15        & 1049  \\
    Baseball                      & BAS    & 16        & 337   \\
    California Housing            & CAL    & 8         & 20640 \\
    MV Artificial Domain          & MV     & 10        & 40768 \\
    House-16H                     & HOU    & 16        & 22784 \\
    Elevators                     & ELV    & 18        & 16559 \\
    Computer Activity             & CA     & 21        & 8192  \\
    Pole Telecommunications       & POLE   & 26        & 14998 \\
    Pumadyn                       & PUM    & 32        & 8192  \\
    Ailerons                      & AIL    & 40        & 13750 \\
    \bottomrule
    \end{tabular}
\caption{The 28 datasets of the experimental study.}
\label{tab:datasets}
\end{table}

In the following subsections we show the results obtained by the different parts of the algorithm.
Moreover, the results obtained by the FRULER algorithm are compared with other state of the art approaches.

\subsection{Experimental Setup}
FRULER was designed to keep the number of parameters as low as possible. 
For the instance selection technique, no parameters are needed.
In the multi-granularity fuzzy discretization, the fuzziness parameter used for the generation of the fuzzy intervals from the split points was $1$, i.e., the highest fuzziness value.
For the evolutionary algorithm, the values of the parameters were: population size $ = 61$, maximum number of evaluations $ = 100,000$, $p_{cross} = 1.0$, $p_{mut} = 0.2$, and $n_{ls} = 5$.
For the generation of the TSK fuzzy rule bases, the weight of the tradeoff between $\ell_1$ and $\ell_2$ regularizations on the Elastic Net is $\alpha = 0.95$, and the regularization parameter $\lambda$ was obtained from a grid search in the interval $[1, 1E-10]$. $\eta^0$ was obtained halving the initial value ($0.1$) until the result worsens.

A 5-fold cross validation was used in all the experiments.
Moreover, 6 trials (with different seeds for the random number generation) of FRULER were executed for each 5-fold cross validation.
Thus, a total of 30 runs were obtained for each dataset.
The results shown in the next section are the mean values over all the runs.
The time measures have been done using a single thread in an Intel Xeon Processor E5-2650L (20M Cache, 1.80 GHz, 8.00 GT/s Intel QPI).

\subsection{Performance of the Instance Selection Process}
We considered two different measures to evaluate the instance selection process:
\begin{itemize}
\item Reduction: is the percentage of reduction in the number of examples, defined as:
\begin{equation}
\textit{Reduction} = \left(1 - \frac{|E_s|}{|E_{tra}|}\right) \cdot 100
\end{equation}
where $|E_s|$ is the number of examples in the subset of selected examples and $|E_{tra}|$ is the original number of examples in the training set.
\item Increase in error: is the increment in the error after applying the instance selection process, defined as:
\begin{equation}
\textit{Increase} = \frac{\epsilon_{E_s}}{\epsilon_{E_{tra}}}
\end{equation}
where $\epsilon_E$ is the mean squared error obtained using \texttt{leave-one-out 1NN} for regression.
\end{itemize}

\begin{table}[htb!]
\centering
    \begin{tabular}{lrS[table-format=8.2]r}
    \toprule
    Data sets & Reduction (\%) & {Increase in error}& Time (m:s) \\
    \midrule
    ELE1       & 83.4     & 0.847             & 00:54     \\
    PLA        & 91.8     & 0.762             & 01:54     \\
    QUA        & 70.9     & 1.073             & 03:50     \\
    ELE2       & 84.9     & 4.075             & 02:09     \\
    FRIE       & 79.5     & 1.529             & 01:33     \\
    MPG6       & 81.6     & 1.239             & 00:45     \\
    DELAIL     & 96.9     & 1.040             & 10:59     \\
    DEE        & 77.7     & 1.139             & 00:42     \\
    DELELV     & 94.0     & 0.968             & 15:44     \\
    ANA        & 98.8     & 16.613            & 05:09     \\
    MPG8       & 81.4     & 0.919             & 00:44     \\
    ABA        & 91.7     & 0.957             & 06:13     \\
    CON        & 88.0     & 1.328             & 02:04     \\
    STP        & 73.6     & 2.351             & 02:07     \\
    WAN        & 85.4     & 1.575             & 02:03     \\
    WIZ        & 64.2     & 1.362             & 02:37     \\
    FOR        & 93.3     & 0.543             & 00:58     \\
    MOR        & 83.4     & 4.277             & 02:12     \\
    TRE        & 81.6     & 5.345             & 02:20     \\
    BAS        & 83.5     & 1.610             & 00:38     \\
    CAL        & 91.6     & 1.270             & 39:44     \\
    MV         & 98.7     & 3.867             & 40:33     \\
    HOU        & 95.4     & 1.118             & 46:08     \\
    ELE        & 96.0     & 1.325             & 30:49     \\
    CA         & 98.9     & 7.396             & 11:57     \\
    POLE       & 98.7     & 18.364            & 27:15     \\
    PUM        & 80.3     & 1.007             & 14:48     \\
    AIL        & 95.4     & 1.118             & 24:31     \\
    \bottomrule
    \end{tabular}
\caption{Average (5-fold cross validation) results obtained by the instance selection for regression method for each dataset.}
\label{tab:isresults}
\end{table}

Table \ref{tab:isresults} shows the average values of reduction and error increase for each data set.
The percentage of reduction achieved is, in general, over 80\% in most of the datasets.
Another four datasets (QUA, FRIE, DEE, STP) have a reduction in the range 70-80\%, and only one dataset (WIZ) has a reduction below 70\% (64.2\%).
The reduction rate does not depend neither on the size of the dataset, nor on the number of variables, but on the complexity of the data.
On the other hand, the increase in 1NN error is very low, as it is greater than 2 for only eight datasets (ELE2, ANA, STP, MOR, TRE, MV, CA, POLE).
The time needed for the execution of the instance selection process is generally low, and only the large scale problems consume more than 15 minutes.

\subsection{Performance of the Multi-Granularity Fuzzy Discretization Process}
We evaluated the discretization with three different measures:
\begin{itemize}
\item Average maximum granularity (over all the variables)
for each dataset:
This measure summarizes the complexity of the fuzzy partitions generated by the discretization.
\item Maximum granularity among the variables for each dataset. 
This represents the upper bound of the fuzzy partitions obtained for each dataset.
It is expected that the smaller this value, the simpler the models obtained by FRULER.
\item The number of variables that have not been discretized at all, i.e., their maximum granularity is equal to 1.
\end{itemize}

\begin{table}[htb!]
\centering
\sisetup{round-precision=1}
    \begin{tabular}{lS[table-format=3.1]S[table-format=3.1]S[table-format=6.1]r}
    \toprule
    Problem & {Average} & { Max}    & {\# Not used} & Time (s:ms) \\
    \midrule
    ELE1     & 2.300   & 2.400  & 0.000       & 00:11      \\
    PLA      & 3.500   & 4.400  & 0.000       & 00:17      \\
    QUA      & 3.933   & 7.800  & 0.000       & 00:26      \\
    ELE2     & 5.100   & 7.600  & 0.000       & 00:21      \\
    FRIE     & 2.000   & 2.000  & 0.000       & 00:18      \\
    MPG6     & 3.560   & 5.800  & 0.000       & 00:13      \\
    DELAIL   & 9.760   & 14.000 & 0.000       & 00:65      \\
    DEE      & 2.167   & 3.000  & 0.000       & 00:11      \\
    DELELV   & 7.800   & 15.400 & 0.800       & 00:63      \\
    ANA      & 1.971   & 6.600  & 5.000       & 00:13      \\
    MPG8     & 2.857   & 5.600  & 1.000       & 00:09      \\
    ABA      & 3.950   & 7.800  & 1.000       & 00:35      \\
    CON      & 5.950   & 14.000 & 1.000       & 00:29      \\
    STP      & 4.267   & 9.400  & 0.000       & 00:24      \\
    WAN      & 5.044   & 14.800 & 0.000       & 00:28      \\
    WIZ      & 4.267   & 9.000  & 0.000       & 00:25      \\
    FOR      & 2.433   & 6.000  & 4.000       & 00:15      \\
    MOR      & 3.947   & 8.000  & 0.000       & 00:27      \\
    TRE      & 3.707   & 6.200  & 0.000       & 00:35      \\
    BAS      & 2.575   & 5.200  & 4.000       & 00:13      \\
    CAL      & 5.125   & 13.800 & 0.000       & 01:40      \\
    MV       & 3.380   & 18.600 & 3.000       & 02:91      \\
    HOU      & 3.675   & 12.200 & 5.000       & 02:20      \\
    ELE      & 8.033   & 17.200 & 2.000       & 01:57      \\
    CA       & 4.143   & 14.400 & 8.000       & 00:73      \\
    POLE     & 4.523   & 16.000 & 5.000       & 01:05      \\
    PUM      & 2.044   & 3.200  & 0.000       & 01:41      \\
    AIL      & 6.685   & 19.0   & 6.2         & 02:01      \\
    \bottomrule
    \end{tabular}
\caption{Average (5-fold cross validation) results obtained by the multi-granularity fuzzy discretization process for each dataset.}
\label{tab:discresults}
\end{table}

Table \ref{tab:discresults} summarizes the results for each dataset.
The average maximum granularity is below 9 in all the cases except for DELAIL dataset.
Moreover, the maximum granularity is always below 20 and only in 11 cases (DELAIL, DELELV, CON, WAN, CAL, MV, HOU, ELE, CA, POLE, AIL) it is above granularity 10.
Even in the datasets with high granularities, the maximum number of fuzzy sets does not generate a huge search space for the evolutionary algorithm.
Finally, the number of variables without discretization is 0 in most of the cases.
In terms of computational time, the discretization module has almost no cost, as the most expensive discretization process is less than 3 seconds.

\subsection{Statistical Analysis}
In this section we compare FRULER with three genetic approaches that are the most accurate genetic fuzzy systems for regression in the literature:
\begin{itemize}
\item FS\textsubscript{MOGFS}\textsuperscript{e}+TUN\textsuperscript{e} \cite{alcala2011fast}: a multi-objective evolutionary algorithm that learns Mamdani fuzzy rule bases. This algorithm learns the granularities from uniform multi-granularity fuzzy partitions (up to granularity $7$) and the lateral displacement of the labels. It includes a post-processing algorithm for tuning the parameters of the membership functions and for rule selection.
\item L-METSK-HD\textsuperscript{e} \cite{gacto2014metsk}: a multi-objective evolutionary algorithm that learns linguistic TSK-0 fuzzy rule bases. The algorithm learns the granularities from uniform multi-granularity fuzzy partitions (up to granularity $7$).
\item A-METSK-HD\textsuperscript{e} \cite{gacto2014metsk}: a multi-objective evolutionary algorithm that learns approximative TSK-1 fuzzy rule bases. The algorithm starts with the solution obtained on the first stage and applies a tuning of the membership functions, rule selection and a Kalman-based calculation of the consequents of the rules.
\end{itemize}

\begin{table*}[htb!]
\centering
    \let\oldtabular\tabular
    \renewcommand{\tabular}{\small\oldtabular}
    \begin{tabular}{lrrrrrrrr}
    \toprule
    algorithms & \multicolumn{2}{c}{FRULER} & \multicolumn{2}{c}{FS\textsubscript{MOGFS}\textsuperscript{e}+TUN\textsuperscript{e}} & \multicolumn{2}{c}{L-METSK-HD\textsuperscript{e}} & \multicolumn{2}{c}{A-METSK-HD\textsuperscript{e}} \\
    \cline{2-9}
    & \# Rules & Test Error & \# Rules & Test Error & \# Rules & Test Error & \# Rules & Test Error\\
    \midrule
    ELE1        & \textbf{4.1}  & 2.012             & 8.1  & \textbf{1.954}  & 15   & 1.925   & 11.4 & 2.022   \\
    PLA         & \textbf{1.4}  & 1.219             & 18.6 & 1.194  & 23   & 1.218   & 19.2 & \textbf{1.136}   \\
    QUA         & 7.8           & 0.0181             & \textbf{3.2}  & \textbf{0.0178}   & 35.9 & 0.019   & 18.3 & 0.0181   \\
    ELE2        & \textbf{4.3}  & 6,729         & 8             & 10,548   & 59   & 20,095   & 36.9 & \textbf{3,192}    \\
    FRIE        & \textbf{8.0}  & \textbf{0.731}    & 22            & 3.138   & 95.1 & 3.084   & 66   & 1.888   \\
    MPG6        & \textbf{13.7} & \textbf{3.727}    & 20            & 4.562   & 99.6 & 4.469   & 53.6 & 4.478   \\
    DELAIL      & \textbf{2.5}  & 1.458             & 6.2           & 1.528   & 98.3 & 1.621   & 36.8 & \textbf{1.402}   \\
    DEE         & \textbf{7.9}  & \textbf{0.080}    & 18.3          & 0.093   & 96.4 & 0.095   & 50.6 & 0.103   \\
    DELELV      & \textbf{5.8}  & 1.045             & 7.9           & 1.086   & 91   & 1.119   & 39.1 & 1.031   \\
    ANA         & \textbf{3.9}  & 0.008             & 10            & \textbf{0.003}   & 48.9 & 0.006   & 33.3 & 0.004   \\
    MPG8        & \textbf{12.7} & \textbf{4.084}    & 23            & 4.747   & 98.7 & 5.61    & 64.2 & 5.391   \\
    ABA         & \textbf{4.5}  & 2.393             & 8             & 2.509   & 42.4 & 2.581   & 23.1 & \textbf{2.392}   \\
    CON         & \textbf{8.9}  & \textbf{20.598}   & 15.4          & 32.977  & 96.5 & 38.394  & 53.7 & 23.885  \\
    STP         & 42.4          & \textbf{0.353}   & \textbf{23}   & 0.912   & 100  & 0.78    & 66.4 & 0.387   \\
    WAN         & \textbf{5.6}  & \textbf{0.888}    & 8             & 1.635   & 91.1 & 1.773   & 48   & 1.189   \\
    WIZ         & \textbf{8.9}  & \textbf{0.663}    & 10            & 1.011   & 55.4 & 1.296   & 29.1 & 0.944   \\
    FOR         & \textbf{5.6}  & \textbf{2,214} & 10            & 2,628    & 93.7 & 4,633    & 40.6 & 5,587    \\
    MOR         & 7.9           & \textbf{0.007}    & 7             & 0.019   & 40.9 & 0.028   & 27.2 & 0.013   \\
    TRE         & \textbf{4.5}  & \textbf{0.027}    & 9             & 0.044   & 42.8 & 0.052   & 28.1 & 0.038   \\
    BAS         & \textbf{6.2}  & 305,777        & 17            & \textbf{261,322}  & 95.7 & 320,133  & 59.8 & 368,820  \\
    CAL         & 15.4          & 2.110             & \textbf{8.4}  & 2.95    & 99.8 & 2.638   & 55.8 & \textbf{1.71}    \\
    MV          & \textbf{6.0}  & 0.083             & 14            & 0.158   & 76.4 & 0.244   & 56.5 & \textbf{0.061}   \\
    HOU         & 12.1          & \textbf{8.005}    & \textbf{11.7} & 9.4     & 68.9 & 10.368  & 30.5 & 8.64    \\
    ELE         & \textbf{5.4}  & \textbf{2.934}    & 8             & 9       & 76.4 & 8.9     & 34.9 & 7.02    \\
    CA          & \textbf{7.1}  & \textbf{4.634}    & 14            & 5.216   & 71.3 & 5.88    & 32.9 & 4.949   \\
    POLE        & 40.8          & 110.898           & \textbf{13.1} & 102.816 & 100  & 150.673 & 46.3 & \textbf{61.018}  \\
    PUM         & \textbf{7.8}  & 0.367             & 17.6          & 0.292   & 87.5 & 0.594   & 63.3 & \textbf{0.287}   \\
    AIL         & \textbf{8.5}  & \textbf{1.404}    & 15            & 2       & 99.1 & 1.822   & 48.4 & 1.51    \\
    \bottomrule
    \end{tabular}
\caption{Average results for the different algorithms.
The test errors in this table should be multiplied by $10^5, 10^{-8}, 10^{-6}, 10^9, 10^8, 10^{-6}, 10^{-4}, 10^{-8}$ in the case of ELE1, DELAIL, DELELV, CAL, HOU, ELV, PUM, AIL respectively.\label{tab:results}}
\end{table*}

Table \ref{tab:results} shows the average results of FRULER and the three algorithms selected for comparison.
Two different results are shown for each algorithm and dataset: the number of rules of the obtained rule base, and the test error measured using equation (\ref{eq:mse}) over the test data.
These indicators allow to compare both the simplicity and the accuracy of the learned models.
The values with the best accuracy ---lowest error--- and best number of rules in table \ref{tab:results} are marked in bold.

It can be seen that the number of rules of FRULER is the lowest in the majority of the datasets.
It should be noted that the number of rules in the large scale problems (the last 8 problems) is also low despite the high number of examples.
Only in 5 problems the FS\textsubscript{MOGFS}\textsuperscript{e}+TUN\textsuperscript{e} Mamdani proposal produces the lowest number of rules.
In the case of accuracy, in 15 of the 28 problems FRULER achieves the best results.
In the other 13 datasets, the best results are for FS\textsubscript{MOGFS}\textsuperscript{e}+TUN\textsuperscript{e} (best in 4 problems) and A-METSK-HD\textsuperscript{e} (best in 9 problems).
From the results, we did not find influence in the performance of FRULER by neither the training dataset size nor the dimensionality of the problem.

In order to analyze the statistical significance of these results the STAC platform \cite{stac} was used to apply the statistical tests. 
A Friedman test was used for both the number of rules and the test error in order to get a ranking of the algorithms and check whether the differences between them were statistically significant.

\begin{table}[htb!]
\centering
    \begin{tabular}{lr}
    \toprule
    Algorithm   &   Ranking \\
    \midrule
    FRULER      &   1.714 \\
    A-METSK-HD\textsuperscript{e}  &   2.036 \\
    FS\textsubscript{MOGFS}\textsuperscript{e}+TUN\textsuperscript{e}     &   2.786 \\
    L-METSK-HD\textsuperscript{e}  &   3.464 \\   
    \midrule
    p-value     & $< 1E-5$ \\
    \bottomrule
    \end{tabular}
\caption{Friedman test ranking results for the test error in table \ref{tab:results}.}
\label{tab:friedman_tst}
\end{table}

\begin{table}[htb!]
\centering
    \begin{tabular}{lrr}
    \toprule
    Comparison              &    p-value  \\
    \midrule 
    FRULER vs A-METSK-HD\textsuperscript{e}         &   0.079 \\
    \bottomrule
    \end{tabular}
\caption{Wilcoxon comparison for the two most accurate algorithms of table \ref{tab:friedman_tst}.}
\label{tab:holm_tst}
\end{table}

Table \ref{tab:friedman_tst} shows the ranking for the test error, with the p-value of the test.
Our proposal ---generates linguistic TSK-1 rules--- gets the top ranking, i.e., it has the best results in accuracy among all the algorithms.
Then, the next algorithm in the ranking is the approximative approach, due to its fine tuning of the rules, followed by the linguistic approaches.
In order to compare whether the difference between FRULER and the second ranked algorithm (A-METSK-HD\textsuperscript{e}\cite{gacto2014metsk}) is significant, a Wilcoxon test was performed (Table \ref{tab:holm_tst}).
The p-value indicates that the difference is statistically significant when using a significance level of $0.1$. 
Thus, even with linguistic rules, FRULER obtains a great accuracy compared to approximative approaches, while getting simpler models.

\begin{table}[htb!]
\centering
    \begin{tabular}{lr}
    \toprule
    Algorithm       & Ranking \\
    \midrule
    FRULER           & 1.214      \\
    FS\textsubscript{MOGFS}\textsuperscript{e}+TUN\textsuperscript{e}       & 1.786 \\
    A-METSK-HD\textsuperscript{e}      & 3 \\
    L-METSK-HD\textsuperscript{e}      & 4 \\
    \midrule
    p-value                                                             & $< 1E-5$ \\
    \bottomrule
    \end{tabular}
\caption{Friedman test ranking results for the number of rules in table \ref{tab:results}.}
\label{tab:friedman_rules}
\end{table}

\begin{table}[ht!]
\centering
    \begin{tabular}{lrr}
    \toprule
    Comparison                                                                       &   p-value  \\
    \midrule 
    FRULER vs A-METSK-HD\textsuperscript{e}       &   $< 1E-4$ \\
    \bottomrule
    \end{tabular}
\caption{Wilcoxon comparison for the two simpler approaches in table \ref{tab:friedman_rules}.}
\label{tab:holm_rules}
\end{table}

To analyze the complexity of the models obtained for each algorithm, the same Friedman test was performed to the number of rules in table \ref{tab:results} (Table \ref{tab:friedman_rules}).
Once again, FRULER has the lowest ranking. The next algorithm in the ranking is the FS\textsubscript{MOGFS}\textsuperscript{e}+TUN\textsuperscript{e} Mamdani approach, followed by the METSK-HD\textsuperscript{e} approaches with a big difference in the ranking.
In order to assess whether the difference in complexity among the most accurate proposals (Table \ref{tab:friedman_tst}) is significant, a Wilcoxon test was also applied (Table \ref{tab:holm_rules}).
The difference is statistically significant (p-value equal to $1E-4$) in the number of rules.
This shows that FRULER obtains accurate solutions with simpler models.

\begin{table}[htb!]
\centering
\small
    \begin{tabular}{l | rrrrrrrrrr}
    \toprule
    Datasets    & ELE1    & PLA     & QUA     & ELE2    & FRIE    & MPG6    & DELAIL  & DEE     & DELELV  & ANA      \\
    \midrule
    Time (h:m:s) & 0:00:51 & 0:01:41 & 0:09:48 & 0:03:05 & 0:05:46 & 0:02:12 & 0:09:58 & 0:02:26 & 0:25:01 & 0:05:05 \\
    Evaluations  & 8,885    & 7,345    & 13,020   & 16,798   & 17,283   & 21,556   & 19,236   & 24,131   & 24,386   & 27,107  \\
    \bottomrule
    \toprule
    Datasets & MPG8    & ABA     & CON     & STP     & WAN     & WIZ     & FOR     & MOR     & TRE     & BAS \\
    \midrule
    Time (h:m:s) & 0:03:29 & 0:17:45 & 0:05:55 & 0:27:41 & 0:10:27 & 0:26:03 & 0:02:28 & 0:27:50 & 0:23:30 & 0:04:41 \\
    Evaluations & 29,355   & 31,537   & 32,318   & 38,468   & 35,812   & 36,168   & 45,367   & 60,101   & 57,569   & 59,362 \\
    \bottomrule
    \toprule
    Datasets & CAL     & MV      & HOU     & ELE     & CA      & POLE    & PUM      & AIL \\
    \midrule
    Time (h:m:s)  & 1:57:03 & 1:17:02 & 4:15:17 & 3:01:30 & 0:38:12 & 1:53:15 & 31:14:27 & 12:50:38 \\
    Evaluations & 33,951   & 35,001   & 61,709   & 68,055   & 78,036   & 99,827   & 96,543    & 100,000 \\
    \bottomrule
    \end{tabular}
\caption{Average run time and number of evaluations per run of FRULER.}
\label{tab:time}
\end{table}

\begin{figure}[htb!]
\centering
\includegraphics[height=0.9\textheight, keepaspectratio]{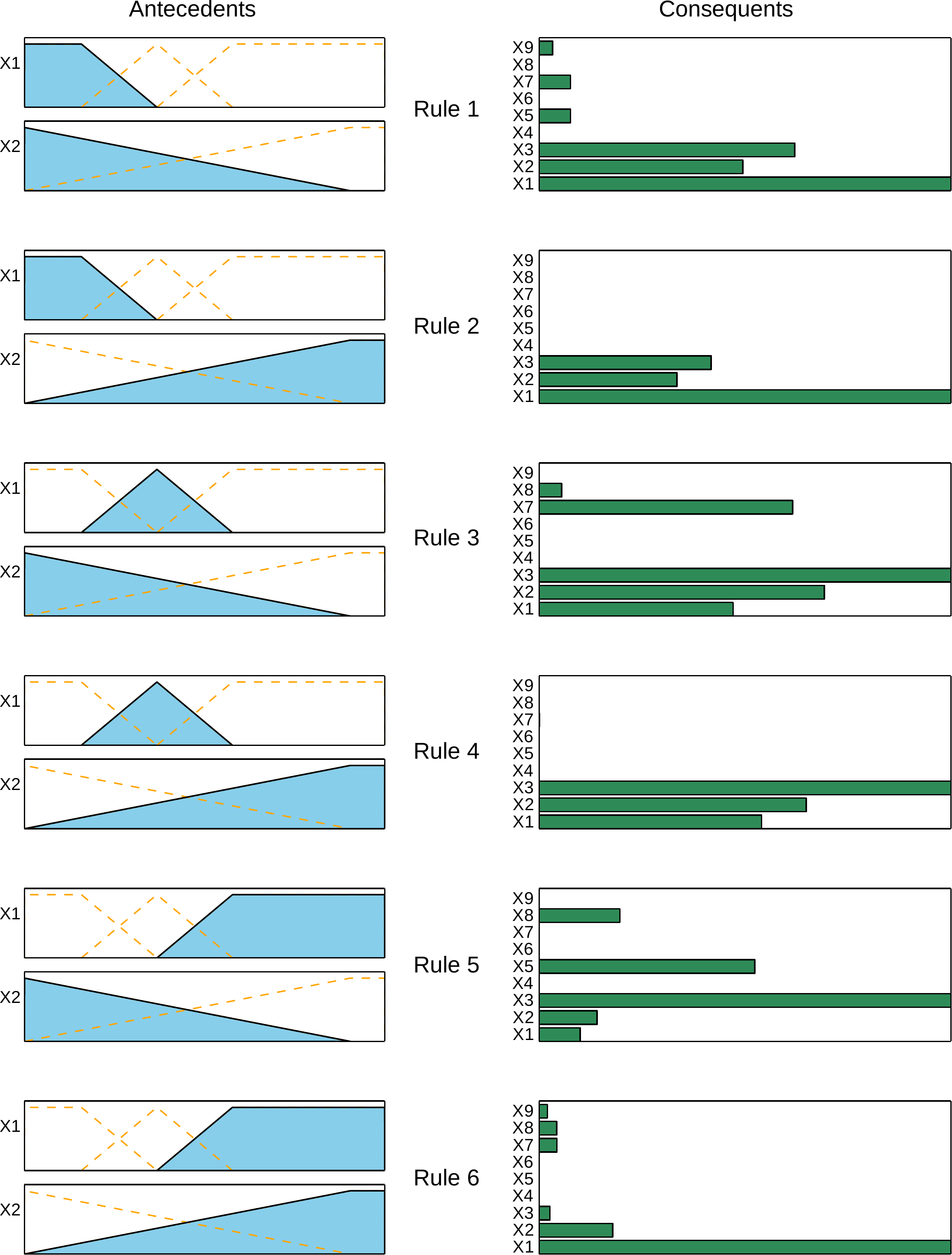}
\caption{An example of TSK fuzzy rule base for the WAN dataset. The system uses only 2 variables for the antecedent part and has 6 different rules. For the sake of simplicity and understandability, the consequents are represented with their absolute value and have been scaled to have the maximum weight equal to 1. 
The test error obtained by this example is 0.885.\label{fig:example}}
\end{figure}

Table \ref{tab:time} shows the average time consumed by a run of FRULER in each dataset.
We also display the number of evaluations until the stopping condition was met.
Although each of the stages of FRULER increases the computational complexity, they contribute to focus the search on the simplest models.
Our method obtains solutions in the range between 1-23 minutes for datasets 1-20 (the most simple ones) and solutions in the range from 1-30 hours for datasets 21-28 (the most complex ones).
Moreover, the number of evaluations is below the $100,000$ limit, except for the largest problem (AIL).
The computational time of FRULER is in the same order of magnitude as A-METSK-HD, being only worse in six datasets (QUA, WIZ, MOR, TRE, PUM and AIL)
\footnote{We do not perform a quantitative comparison with the computational times of \cite{gacto2014metsk} as the processor is not the same}.

In order to demonstrate the simplicity of the models generated by FRULER, Figure \ref{fig:example} shows an example of one of the rule bases generated for the WAN dataset.
The Figure shows two columns for each rule: the fuzzy sets used in the antecedent and the weight of the variables in the consequent.
For the sake of simplicity and understandability, the consequents are represented with their absolute value and have been scaled to put the maximum weight equal to 1.
The antecedent only uses two variables with granularity 3 and 2 respectively, thus 6 rules are needed to cover all the combinations.
On the other hand, the consequent column shows the importance of each input variable for each rule, providing a qualitative understanding of the model.
In this case, the first three variables (X1, X2 and X3) have the greatest importance in the consequent.
Note that, even though this is one of the simplest models obtained by FRULER, the test error is very low (0.885).

\section{Conclusions\label{sec:conclusion}}
In this paper, a novel genetic fuzzy system called FRULER was presented.
FRULER learns simple and linguistic TSK-1 knowledge bases for regression problems.
This new approach has two general-purpose preprocessing stages for regression problems: a new instance selection for regression and a novel non-uniform multi-granularity fuzzy discretization.
The evolutionary learning algorithm incorporates an automatic generation of the TSK fuzzy rule bases from fuzzy partitions that uses Elastic Net in order to obtain consequents with low overfitting.

FRULER was compared with three state of the art algorithms that learn different types of fuzzy rules: linguistic Mamdani, linguistic TSK-0 and approximative TSK-1.
The results were analyzed using statistical tests, which show that FRULER obtains high accuracy, but with a lower number of rules and with a linguistic data base.
This is of particular interest in problems where both high accuracy and interpretability are demanded, in order to provide qualitative understanding of the model to the users.

\section*{Acknowledgments}
This work was supported by the Spanish Ministry of Economy and Competitiveness under projects TIN2011-22935, TIN2011-29827-C02-02 and TIN2014-56633-C3-1-R, and the Galician Ministry of Education under the projects EM2014/012 and CN2012/151. 
I. Rodriguez-Fdez is supported by the Spanish Ministry of Education, under the FPU national plan (AP2010-0627).

\section*{\refname}
\bibliographystyle{elsarticle-harv}
\bibliography{main}

\end{document}